\documentclass[runningheads]{llncs}
\usepackage{graphicx}
\usepackage{multirow}
\usepackage{xcolor}

\begin{document}
\title{A Language-based solution to enable Metaverse Retrieval}
%
%
\author{Ali Abdari\inst{1,2}\orcidID{0000-0002-4482-0479} \and
Alex Falcon\inst{1}\orcidID{0000-0002-6325-9066} \and
Giuseppe Serra\inst{1}\orcidID{0000-0002-4269-4501}}
\authorrunning{F. Author et al.}
%
\institute{University of Udine, Udine, Italy \and University of Naples Federico II, Naples, Italy}
%
\maketitle              
\begin{abstract}
Recently, the Metaverse is becoming increasingly attractive, with millions of users accessing the many available virtual worlds. However, how do users find the one Metaverse which best fits their current interests? So far, the search process is mostly done by word of mouth, or by advertisement on technology-oriented websites. However, the lack of search engines similar to those available for other multimedia formats (e.g., YouTube for videos) is showing its limitations, since it is often cumbersome to find a Metaverse based on some specific interests using the available methods, while also making it difficult to discover user-created ones which lack strong advertisement. To address this limitation, we propose to use language to naturally describe the desired contents of the Metaverse a user wishes to find. Second, we highlight that, differently from more conventional 3D scenes, Metaverse scenarios represent a more complex data format since they often contain one or more types of multimedia which influence the relevance of the scenario itself to a user query. Therefore, in this work, we create a novel task, called Text-to-Metaverse retrieval, which aims at modeling these aspects while also taking the cross-modal relations with the textual data into account. Since we are the first ones to tackle this problem, we also collect a dataset of 33000 Metaverses, each of which consists of a 3D scene enriched with multimedia content. Finally, we design and implement a deep learning framework based on contrastive learning, resulting in a thorough experimental setup.

\keywords{Multimedia \and Text-to-Multimedia Retrieval \and Cross-modal understanding \and Metaverse \and Contrastive Learning}
\end{abstract}

\section{Introduction}

Nowadays, the Metaverse is becoming an increasingly popular pastime where the users relax using a multitude of different applications, mostly revolving around entertainment, including social-oriented games (e.g., Decentraland\footnote{https://decentraland.org/} and Roblox\footnote{https://www.roblox.com/}), fitness activity (e.g., ''The thrill of the fight'' and ''Archer VR'', two virtual reality applications for boxing and archery, respectively), and much more. Notably, the popularity of the Metaverse is ever-increasing and so far has reached a total of around 400 million users logging in monthly \cite{metaversed2023users}. As an example, the amount of users logging in daily into the Roblox has grown from around 10 million at the end of 2018 to a staggering 70 million during the first part of 2023 \cite{statista2020roblox}. Nonetheless, the Metaverse is not all about entertainment: in fact, many Metaverses are being created for more professional use cases, for instance for industrial training \cite{almeida2023vrtraining} or predictive maintenance applications \cite{agnusdei2021classification,siyaev2021towards}. 

\begin{figure}[t]
    \centering
    \includegraphics[width=\linewidth,trim=0cm 14cm 0cm 0cm,clip]{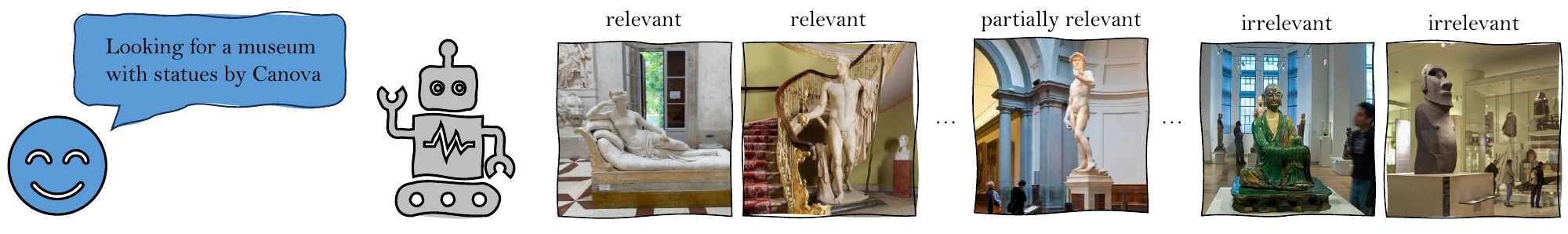}
    \caption{An illustration of the Text-to-Metaverse retrieval task, which requires the model to rank the available Metaverses by estimating their relevance to a user-defined query.}
    \label{fig:problem}
\end{figure}

The process of discovering a Metaverse that best matches one's interests is unfortunately not easy: the current means for discovering Metaverses are word-of-mouth (e.g., on social media, dedicated forums, etc.) and advertisement in technology-oriented websites. Therefore, similarly to what happened with other formats of digital content, there is a need for the user to express their interests through a natural language query. For instance, this happened with Google Images for visual data, Instagram or TikTok for videos, and YouTube for audio. To address the lack of research interest in this topic, in this paper, we propose a novel task, called Text-to-Metaverse retrieval, whose objective is to rank a list of Metaverses based on their relevance to a user-defined query, as illustrated in Figure \ref{fig:problem}. In this way, we empower the users and give them more freedom in choosing how they want to express their interests. 

A second point to note is that Metaverses have greater diversity and dynamism than conventional 3D scenes. In fact, Metaverses contain one or more types of multimedia data. For instance, a user could travel millions of years back in time within a Metaverse and interact with digital reconstructions of extinct animals, whereas a virtual shopping outlet could be hosted in a Metaverse so users can try on different clothes. Interestingly, multimedia content plays a fundamental role in defining the relevance of the Metaverse to the user query within the search engine. Moreover, their multimedia content can also change over time. For instance, the exhibition hosted at a Metaverse museum could be temporary and therefore it could significantly change from time to time. Unfortunately, there are two major issues with the public datasets available for 3D scenes: first, they do not contain multiple instances of the same Metaverse at different times; second, their scenes do not present any multimedia content. Therefore, to benchmark the progress on the Text-to-Metaverse retrieval task, we collect a dataset composed of around 33000 Metaverses, each of which contains a painting, hence focusing our work on art-related Metaverses.

Finally, we design and implement several solutions based on recent advancements in deep learning and cross-modal retrieval. Our results show that a solution to the Text-to-Metaverse retrieval is feasible and may provide a great tool for users to filter irrelevant Metaverses, further motivating the research on this very important and emergent topic.

The main contributions of our work can be summarized as follows:
\begin{itemize}
    \item We introduce and motivate the Text-to-Metaverse retrieval task, which is becoming an increasingly important task for Metaverse understanding and exploration.
    \item We collect a dataset of around 33000 art-related Metaverses, and annotate each of them with a textual description of its contents.
    \item We design and implement a framework comprising several solutions based on deep learning, which experimentally motivates the feasibility of solutions for this important topic.
\end{itemize}

The rest of the paper is organized as follows. In Section \ref{sec:rw}, we review the literature on related topics, highlighting the lack of datasets on Metaverses and the need for further work on this topic. In Section \ref{sec:proposed}, we describe in detail our methodology, which is based on recent advancements in 3D scene understanding and cross-modal retrieval. Section \ref{sec:exps} reports the experimental results observed on our dataset. Finally, in Section \ref{sec:conclusions}, we highlight future work and draw conclusions on our research study.

\section{Related work\label{sec:rw}}

In this Section, we explore the literature related to the proposed task, Text-to-Metaverse retrieval. First, in Section \ref{sec:rw_metaverse} we review the literature on Metaverse-related technologies, to contextualize the work and highlight the lack of research directions in text-guided retrieval for Metaverses. Second, the area on cross-modal retrieval has greatly grown over the past few years and gives us access to a vast knowledge related to how to perform text-guided retrieval (Section \ref{sec:rw_cross}). 

\subsection{Background on Metaverse-related research\label{sec:rw_metaverse}}

The term Metaverse was first mentioned in the sci-fi novel ''Snow Crash'' by Neal Stephenson, who envisioned it as the successor of the Internet and as a form of multiplayer online game with user-controlled digital avatars. More recently, Lee et al. defined it as a virtual world where synthetic elements are blended into the real world \cite{lee2021all}. Nowadays, extended reality technologies are used to enable users to access the Metaverse, and new devices are being presented frequently as the technology for it becomes more accessible to customers. Along the same track, new use cases are being developed at an increasing rate, resulting in many applications ranging from entertainment, such as digital museums \cite{choi2017content,lee2022proposal} and concerts \cite{jin2022metamgc}, passing through smart healthcare and assistance \cite{laaki2019prototyping,liu2019novel}, to industry-ready applications, such digital shopping and virtual try-on \cite{dawson2022data,song2022vtonshoes} to predictive maintenance on industrial equipment \cite{agnusdei2021classification,siyaev2021towards}. Notably, these applications are often supported by foundational research being performed on related problems, such as human pose estimation \cite{wang2020avatarmeeting,zhou2023metafipp} and content generation \cite{wang2023intelligent}.

However, the task of searching for a Metaverse based on user interests has not been addressed in the literature so far, highlighting the importance of this new research direction.

\subsection{Cross-modal understanding and retrieval applications\label{sec:rw_cross}}

The process of finding query-relevant content in a large dataset is commonly regarded as a query-guided retrieval problem. During the past few years, multimedia retrieval has become increasingly popular due to the large increase in user-generated content on social media and the need to filter irrelevant content for a specific user. Specifically, considerable advancements were obtained in text-guided retrieval for videos \cite{2023clip4clip,ge2022bridging}, images \cite{cheng2022cross,radford2021learning}, and audio \cite{wu2023large,xin2023improving}, mostly thanks to the surge of deep representations. A common way to learn neural models for cross-modal retrieval consists in the use of contrastive loss functions to automatically learn how to address the domain gap \cite{miech2020end,schroff2015facenet}. In fact, contrastive loss functions aim at maximizing the similarity of paired samples in the dataset, e.g., an image and its caption, allowing for the subsequent use of the learned functions to map the queries into the same embedding space and perform retrieval.

Interestingly, the 3D scene retrieval problem, which could be seen as greatly related to the proposed Text-to-Metaverse retrieval, has not been addressed in the literature so far. In fact, the research on this topic is often focused on specific aspects of the scene, such as identifying the text \cite{wang2021scene,wen2023visual} or the objects present in the scenes \cite{nguyen2020robot}, or retrieving 3D scenes using 2d images \cite{abdul2018shrec,abdul2019shrec} or sketches as the input queries \cite{yuan2019sketch}. In particular, differently from the SHREC challenge \cite{abdul2018shrec,abdul2019shrec}, the Text-to-Metaverse retrieval task requires free-form text queries and the Metaverses contain multimedia content which is not available in the 3D scenes used in SHREC. Therefore, the datasets available for the aforementioned tasks are not usable for our task, raising the necessity to collect a dataset for it.


\section{Proposed methodology\label{sec:proposed}}

In this Section, we describe the three main components of our methodology. First, in Section \ref{sec:proposed_arch}, we provide a thorough description of the network architecture used in all our experiments. Second, Section \ref{sec:proposed_dataset} describes the steps we performed to collect the dataset used in the experiments.

\begin{figure}
    \centering
    \includegraphics[width=\linewidth,trim=0.2cm 7.25cm 1.1cm 0.25cm,clip]{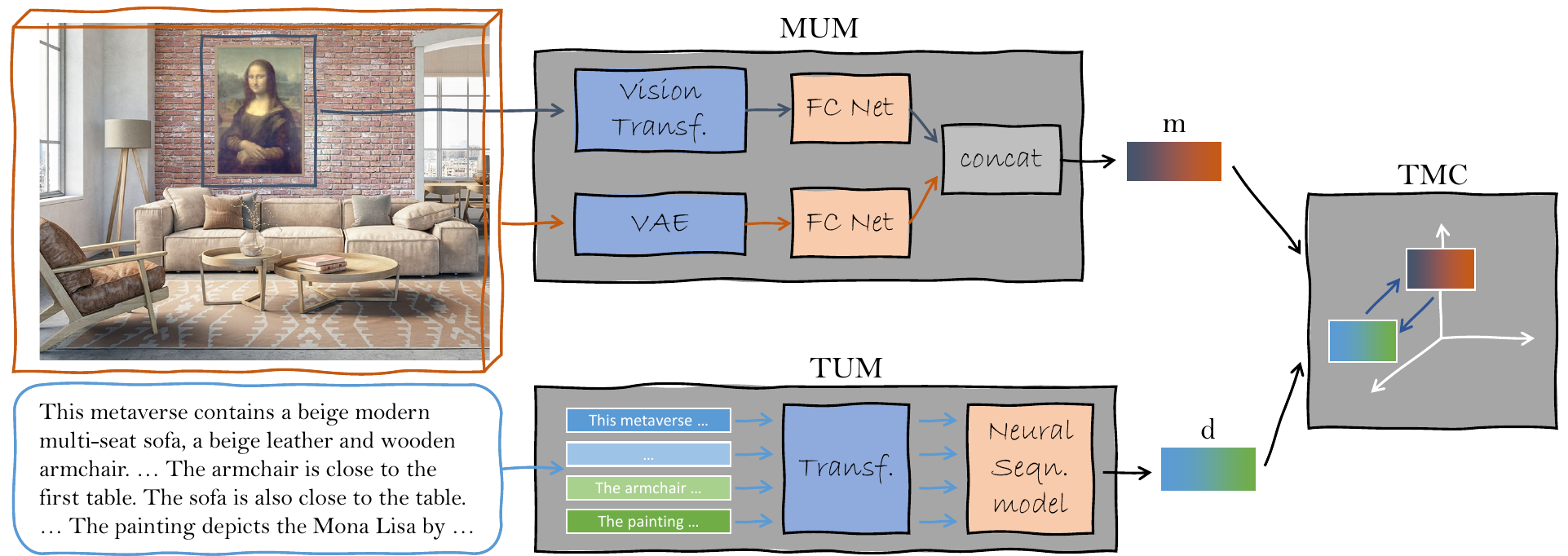}
    \caption{Overview of the network architecture used in the experiments. A thorough discussion is present in Section \ref{sec:proposed_arch}.}
    \label{fig:net_arch}
\end{figure}

\subsection{Network architecture\label{sec:proposed_arch}}
Figure \ref{fig:net_arch} presents an overview of the proposed architecture. As can be seen, it is composed of three main components: MUM, the Metaverse Understanding Module; TUM, the Textual Understanding Module; and, finally, the Text-Metaverse Contrastive learning framework (TMC). All of these components are described in detail in the rest of this section.

\subsubsection{MUM: Metaverse Understanding Module}
As the name suggests, the MUM is used to model the Metaverse scenarios under analysis. It is assumed that the paintings can be extracted from the room. Then, the paintings are initially processed separately from the scene. Specifically, the representation of the scene $\rho_s$ is obtained by using a recent Variational Autoencoder \cite{yang2021scene}, whereas a Vision Transformer trained with CLIP is used for the painting representation $\rho_p$ \cite{radford2021learning}. Then, a deep network, FCNet, is used to learn two separate representations of the inputs, using a similarly structured architecture:
\begin{eqnarray}
    r_1 = \mathrm{ReLU}(W_1 \rho_\star + b_1)\\
    r_2 = \mathrm{ReLU}(W_2 \, \mathrm{BatchNorm}(\delta_1(r_1))) + b_2\\
    r_\star = W_3 \, \mathrm{BatchNorm}(\delta_2(r_2))) + b_3
\end{eqnarray}
\noindent where $\rho_\star$ is either $\rho_s$ or $\rho_p$, and $\delta$ is the dropout operator. The output of this FCNet is $r_s \in R^{1 \times \frac{H}{2}}$ when the input is $\rho_s$, and $r_p \in R^{1 \times \frac{H}{2}}$ when the input is $\rho_p$. Finally, the output of MUM is a vectorial representation $m$ of size $1 \times H$ obtained by concatenating $r_s$ and $r_p$.

\subsubsection{TUM: Textual Understanding Module}
To obtain the representation of the textual data, TUM consisting of the following steps is employed. First, the Metaverse descriptions are separated into sentences by using the period as a splitting term. Then, a representation of each sentence is obtained by using the same CLIP module used in the previous section. This process is used since the descriptions are very long, hence difficult to model with standard natural language processing techniques (e.g., BERT has a context of 512 tokens \cite{devlin2019google}). After obtaining the sentence-level representations, a contextual representation for the full description is obtained through a neural sequence model (more details in Section \ref{sec:exps}). Ultimately, the final representation $d \in R^{1 \times H}$ for the description is given by the last hidden state of the sequence model.

\subsubsection{TMC: Text-Metaverse Contrastive learning}
The final step of our methodology is given by the Text-Metaverse contrastive learning framework, TMC. We use this approach because of its great success in recent years in cross-modal retrieval applications \cite{radford2021learning}. Given a batch of B Metaverses and their descriptions, the respective representations are obtained through MUM and TUM. Then, the similarity between a Metaverse $m_i$ and its description $d_i$ is maximized through a contrastive loss function. Specifically, the standard triplet loss \cite{schroff2015facenet} is used, which is described by the following equations:
\begin{eqnarray}
    \mathcal{L}_{mt, i, j} = relu(0, \Delta + sim(s_i, d_j) - sim(s_i, d_i))\\
    \mathcal{L}_{tm, i, j} = relu(0, \Delta + sim(s_j, d_i) - sim(s_i, d_i))\\
    \mathcal{L} = \frac{1}{B} \sum_i \sum_{j \ne i} \mathcal{L}_{tm, i, j} + \mathcal{L}_{mt, i, j}
\end{eqnarray}
\noindent where $\Delta$ is a fixed margin, and $\mathcal{L}$ is the final loss used to perform the training.

\subsection{Dataset collection\label{sec:proposed_dataset}}
As mentioned in the previous sections, Metaverses can be seen as 3D scenes in which there is additional multimedia content playing a fundamental role in defining the purpose of the Metaverse itself. However, as mentioned in Section \ref{sec:rw_cross}, there are no suitable public datasets which contain multimedia-enriched 3D scenes. Therefore, in the following, we describe the process followed to collect the dataset.

\subsubsection{Collection of suitable Metaverse scenarios}

To create the Metaverses suitable for our task, we started from 3384 indoor scenarios designed by professionals \cite{fu20213d} to provide high-quality 3D scenes. These scenarios feature several types of furniture, including lamps, beds, and wardrobes, among others. Then, inspired by virtual Metaverses covering museums and similar art exhibitions, we decided to make a selection of famous paintings and put those into the 3D scenes to form the final Metaverses. Specifically, we selected ten paintings and then paired each of them to each scenario, resulting in a total number of 33840 Metaverse scenarios.

\subsubsection{Textual descriptions}

To obtain the descriptions for our Metaverses, we devised an automatic procedure consisting of two main phases.

First, given a scenario, the description is obtained by putting together the automatically created sentences, which are separately created for each furniture piece. These are obtained by gathering a set of tags from the 3d-front dataset, and then by putting them together via a template. The tags are divided into categories (e.g., ''Wardrobe'', ''Bed''), style (e.g., ''Chinese'', ''Industrial''), theme (e.g., ''Wrought Iron'', ''Texture Mark'), and material tags (e.g., ''Marble'', ''Cloth''). The template for the first furniture is the following: ''This room contains $<$N$>$ $<$Category$>$ with $<$Style$>$ style, $<$Theme$>$ theme, and $<$Material$>$.'', where N identifies the quantity of repetitions for a given instance, and the other is the tags for that furniture piece. Then, for the other pieces, it slightly changes to: ''Additionally, it also contains $<$N$>$ $<$Category$>$ with $<$Style$>$ style, $<$Theme$>$ theme, and $<$Material$>$.''. After that, a similar procedure is followed to describe the position of the furniture with respect to other furniture pieces. These positional sentences use a template, ''The $<$N$>$ $<$Category1$>$ with $<$Style1$>$ style, $<$Theme1$>$ theme, and $<$Material1$>$ is $<$Distance$>$ from $<$N$>$ $<$Category2$>$ with $<$Style2$>$ style, $<$Theme2$>$ theme, and $<$Material2$>$.'', to provide intuition on the distance separating two objects in the scenario. The words we use for the distance (''so close'', ''close'', ''far'', and ''so far'') are determined by the standard deviation of the distribution of distances.

The second step involves the description of the paintings. After describing the scenario, the description of the painting is created using the following template: ''Also, in this room there is a painting called $<$Name$>$ by $<$Author$>$, which $<$Description$>$.'', where the description is obtained by asking ChatGPT with the following prompt: ''provide two sentences for this painting: $<$Name$>$ by $<$Author$>$'' and manually adapted.

Finally, given a scenario and a painting, we put together their descriptions to obtain the final one. On average, the descriptions are 626.91 tokens long (ranging from 162 to 2014), with an average of 23.33 sentences (ranging from 7 to 79).

\section{Experimental results\label{sec:exps}}

In this section, several experiments are performed to analyze the behavior of the model when architectural changes are applied to MUM (Sec. \ref{sec:exps_mum}) or TUM (Sec. \ref{sec:exps_tum}), followed by a discussion of the results and the limitations of our methodology. To perform the evaluation on the Text-to-Metaverse retrieval task, standard metrics are used: the recall rates at K (R@K, with K=1,5,10,50,100), the median rank (MedR), and the mean rank (MR). All the code and data are available at https://github.com/aliabdari/NLP\_to\_rank\_artistic\_Metaverses . 

\subsection{How to model the Metaverses?\label{sec:exps_mum}}
To model the Metaverses under analysis, we consider an alternative to the proposed MUM. The proposed model can be identified as a late fusion approach (LF), since the representation for the scene and the painting are first treated separately, and then fused. Therefore, here we explore an early fusion approach (EF). Specifically, the CLIP representation of the painting and that of the scene are early fused through concatenation, after learning a linear transformation from 512 CLIP features to D features (same as the scene representation). Then, their concatenation is passed through a FCNet (Sec. \ref{sec:proposed}) to learn a joint representation. The results of the comparison are shown in Table \ref{tab:cmp_mum}. Two major observations can be made. First, both the solutions achieve very high R@100 (74.7\% and 92.1\%). However, it does not mean that the problem is solved since retrieving the most suitable Metaverse among the top 100 Metaverses is far from perfect from the user's perspective. On the other hand, a R@1 of 1.8\% and 15.3\% shows that there is a lot of room for improvement on both the early and late fusion approaches. Second, the late fusion approach achieves far better performance in our experimental setup. This may be due to the great domain gap between the two modalities (scene and painting) which makes it difficult to obtain a meaningful representation by early concatenating the two separate representations.

\begin{table}[t]
    \centering
    \caption{Comparison across different alternatives for our Metaverse understanding module. Details in Section \ref{sec:exps_mum}.}
    \begin{tabular}{c|ccccc|cc|}
        Method & R@1 & R@5 & R@10 & R@50 & R@100 & MedR & MR \\ \hline
        EF &1.8 &5.1 &17.9 &55.6 &74.7 &42.0 &82.4 \\
        LF &\textbf{15.3} &\textbf{41.1} &\textbf{55.6} &\textbf{84.4} &\textbf{92.1} &\textbf{8.0} &\textbf{37.7} 
    \end{tabular}
    \label{tab:cmp_mum}
\end{table}

\subsection{How to model the descriptions?\label{sec:exps_tum}}

Similar to the previous analysis, here we aim to explore different variations for the neural sequence model used in TUM. Specifically, we explore standard methods: temporal average pooling (''Mean''), LSTM, Bidirectional LSTM (''BiLSTM''), GRU, and Bidirectional GRU (''BiGRU''). The results are reported in Table \ref{tab:cmp_tum}. Interestingly, the results show that pooling the sentence representations achieves better performance than using a standard LSTM (e.g., Mean achieves 24.8\% R@5, whereas the LSTM leads to 16.3\%). On the other hand, the GRU-based methods perform better than their LSTM counterpart and, in particular, the BiGRU achieves the best overall performance (15.3\% R@1 compared to 13.2\% obtained by BiLSTM). These results may be explained by the fact that the LSTM has more parameters than the GRU, which may require more diverse data to outperform the simpler GRU architecture. Second, in both cases, bidirectionality is fundamental to achieve a better contextual understanding.

\begin{table}[t]
    \centering
    \caption{Comparison across different alternatives for our textual understanding module. Details in Section \ref{sec:exps_tum}.}
    \begin{tabular}{c|ccccc|cc|}
        Method & R@1 & R@5 & R@10 & R@50 & R@100 & MedR & MR \\ \hline
        Mean &7.5 &24.8 &37.2 &74.5 &86.9 &19.0 &52.8 \\
        GRU &10.8 &35.9 &51.8 &83.2 &90.9 &10.0 &40.6 \\
        BiGRU &\textbf{15.3} &\textbf{41.1} &\textbf{55.6} &\textbf{84.4} &\textbf{92.1} &\textbf{8.0} &\textbf{37.7} \\
        LSTM &4.1 &16.3 &28.6 &65.7 &79.5 &27.0 &96.0 \\
        BiLSTM &13.2 &37.1 &52.2 &82.4 &90.7 &10.0 &43.6 \\ 
    \end{tabular}
    \label{tab:cmp_tum}
\end{table}

\subsection{Limitations}
In this section, the limitations of our work are highlighted.

\textbf{About the Metaverse modeling.} Our Metaverse Understanding Module uses the recent advancements in scene understanding achieved by Yang et al. \cite{yang2021scene} to model the scenario. A different approach is based on large pretrained vision-language models, e.g., CLIP \cite{radford2021learning}. Different from the scene understanding technique, it would allow for improved spatial reasoning, hence enabling a better understanding of the relations between objects and furniture. Moreover, the language-supported training also reduces the domain gap between the two modalities, possibly leading to a smoother training behavior. This possibility is a promising future direction for our work.

\textbf{About the multimedia content.} In our work, the focus is on painting-related Metaverses. However, there are other forms of art, such as mosaics, videos, and 3D artifacts (statues, jewels, armors, etc). Moreover, Metaverses could also contain multimedia content which is not artistic per se, such as game-like or digital traveling experiences. Therefore, future work on this topic should also strive to include and model other media formats.

\subsection{Implementation details}

In our experiments, we used B=64, D=200, H=256. Specifically, the value for D was given by the previous work on scene understanding which was used to get the scene representation \cite{yang2021scene}. To optimize the parameters, we used Adam with an initial learning rate of 0.008, which was reduced by 25\% after 17 epochs. The available data is split into three sets with a 70/15/15 ratio, and we ensure that each scene is not present in multiple sets at the same time. The training lasts 30 epochs, and $\Delta=0.25$ in our loss function. The performance on the testing set is done using the best model on the validation set. To perform the experiments, we used PyTorch 1.12.1 on a machine equipped with an RTX A5000 GPU, an Intel Xeon E5-1620, and 16 GB of RAM.

\section{Conclusions\label{sec:conclusions}}
The Metaverse is becoming increasingly popular among the users for daily use. However, filtering irrelevant Metaverses based on a user query is a difficult task which has not been addressed by research so far. Therefore, in this paper, we introduced the Text-to-Metaverse retrieval task, inspired by the recent advancements in multimedia retrieval. Since the public datasets do not have multimedia-rich 3D scenes, we collected a dataset of 33840 Metaverses, each paired with a detailed textual description. We compared the proposed methodology to several alternatives both for the Metaverse and the textual modeling modules. Finally, we highlighted the limitations of our work which may serve as inspiration for promising research directions.

\section*{Acknowledgments}
This work was supported by the Department Strategic Plan (PSD) of the University of Udine–Interdepartmental Project on Artificial Intelligence (2020-25), MUR Progetti di Ricerca di Rilevante Interesse Nazionale (PRIN) 2022 (project code 2022YTE579), and by TechStar Srl, Italy. Also, we thank Beatrice Portelli for helping with the illustrations and for the useful feedback during the preparation of this work.

\bibliographystyle{splncs04}
\bibliography{biblio}
\end{document}